\documentclass[letterpaper]{article} 
\usepackage{aaai23}  
\usepackage{times}  
\usepackage{helvet}  
\usepackage{courier}  
\usepackage[hyphens]{url}  
\usepackage{graphicx} 
\usepackage{natbib}  
\usepackage{caption} 
\usepackage{algorithm}
\usepackage{algorithmic}

\usepackage{amsmath}
\newcommand{\y}{\mathbf{y}}
\newcommand{\bsigma}{\mathbf{\sigma}}
\newcommand{\bomega}{\boldsymbol{\omega}}
\newcommand{\bL}{\mathbf{L}}
\newcommand{\D}{\mathbf{D}}
\newcommand{\A}{\mathbf{A}}

\usepackage{siunitx}
\usepackage{newfloat}
\usepackage{listings}
\DeclareCaptionStyle{ruled}{labelfont=normalfont,labelsep=colon,strut=off} 
\floatstyle{ruled}
\newfloat{listing}{tb}{lst}{}
\floatname{listing}{Listing}
\usepackage{color} 

\title{Cosmic Microwave Background Recovery: \\ A Graph-Based Bayesian Convolutional Network Approach}
\author{
    Jadie Adams\textsuperscript{\rm 1,2},
    Steven Lu\textsuperscript{\rm 1},
    Krzysztof M. Gorski\textsuperscript{\rm 1},
    Gra{\c c}a Rocha\textsuperscript{\rm 1},
    Kiri L. Wagstaff\textsuperscript{\rm 1}
}
\affiliations{
    \textsuperscript{\rm 1} Jet Propulsion Laboratory, California
    Institute of Technology, 4800 Oak Grove Drive, Pasadena, CA
    91109-8099, USA \\ 
    \textsuperscript{\rm 2} Scientific Computing and Imaging Institute, University of Utah, 201 
    Presidents' Cir, Salt Lake City, UT 84112, USA \\ 
    jadie@sci.utah.edu, \{you.lu,krzysztof.m.gorski,graca.m.rocha\}@jpl.nasa.gov, wkiri@wkiri.com
}

\usepackage{bibentry}

\begin{document}

\maketitle

\begin{abstract} 
The cosmic microwave background (CMB) is a significant source of knowledge about the origin and evolution of our universe. However, observations of the CMB are contaminated by foreground emissions, obscuring the CMB signal and reducing its efficacy in constraining cosmological parameters. We employ deep learning as a data-driven approach to CMB cleaning from multi-frequency full-sky maps. In particular, we develop a graph-based Bayesian convolutional neural network based on the U-Net architecture that predicts cleaned CMB with pixel-wise uncertainty estimates. We demonstrate the potential of this technique on realistic simulated data based on the Planck mission. We show that our model accurately recovers the cleaned CMB sky map and resulting angular power spectrum while identifying regions of uncertainty. Finally, we discuss the current challenges and the path forward for deploying our model for CMB recovery on real observations. 
\end{abstract}

\section{Introduction} 
\begin{figure}[ht!]
    \includegraphics[width=\linewidth]{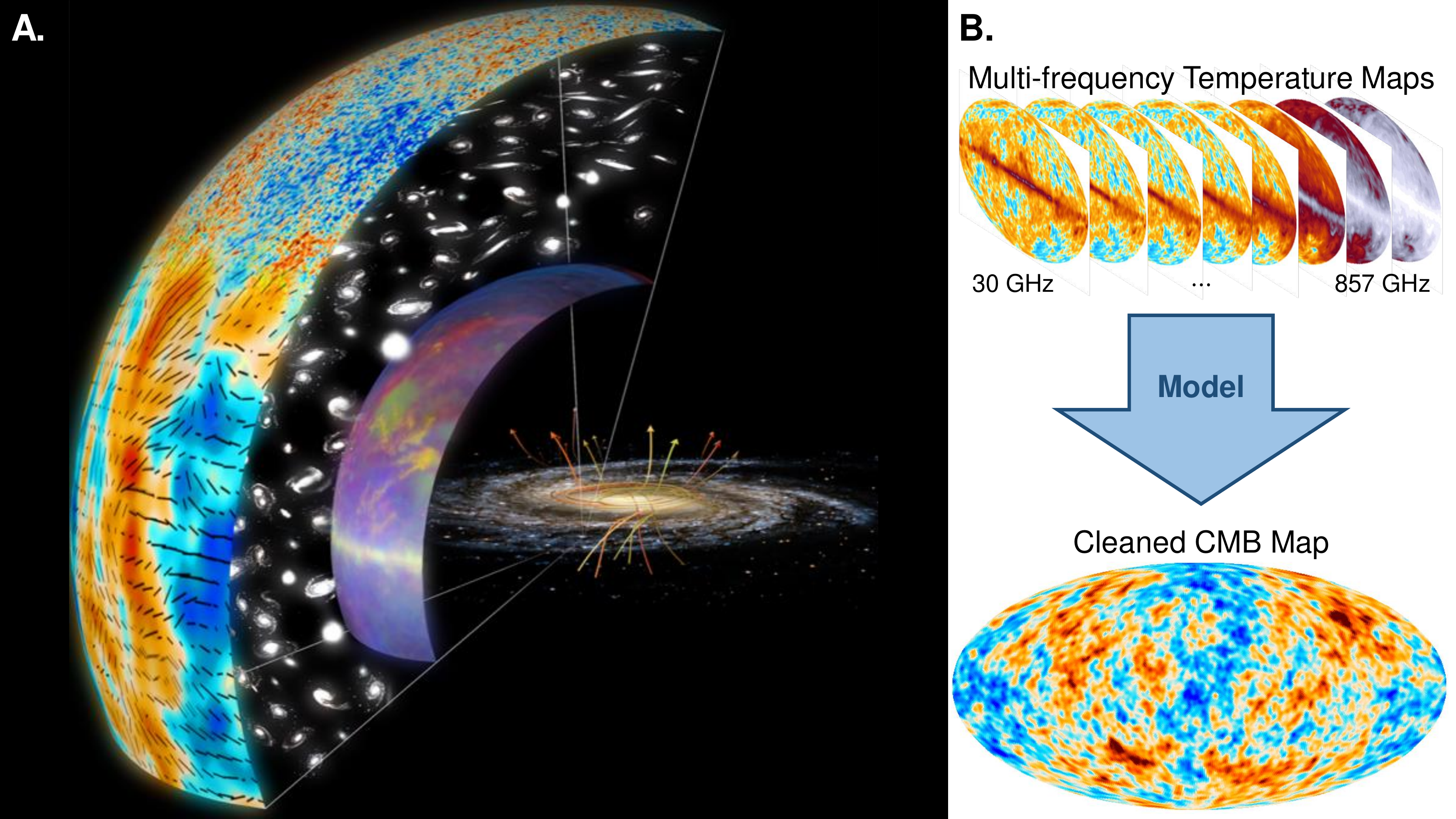}
      \caption{(A) The CMB. This graphic displays our galaxy at the center (including magnetic fields), the intergalactic medium, and the CMB in the outer shell. The color map shows temperature fluctuation and the vector lines represent the polarization amplitude and direction of the radiation. The inner shell displays foreground emissions which block telescopes from observing the CMB. Figure from a Keck Institute for Space Studies (KISS) workshop report \cite{Graca2019KISS}.
      (B) CMB Cleaning Overview. CMB cleaning models take multi-frequency full-sky temperature maps as input and recover the cleaned CMB map.}
      \label{fig:CMB}
\end{figure}

Cosmic Microwave Background (CMB) is remnant electromagnetic radiation from the early universe and a fundamental probe of cosmology. Small anisotropies, or irregularities, in the temperature fluctuations and polarization of CMB sky maps, can constrain models of structure formation and fundamental physics. However, such anisotropies are hidden by foreground emissions (i.e., contamination from Galactic emission, dust emission, synchrotron emission, etc.) as shown in Figure \ref{fig:CMB}. Thus correctly discriminating between CMB and foreground emissions via observations made at multiple frequencies is crucial to cosmological parameter estimation. 

Recent success in applying machine learning techniques to astrophysical tasks has motivated using artificial neural networks to remove foreground and reduce systematic errors in CMB cleaning. 
For such techniques to be maximally valuable for cosmological study, capturing uncertainty or an estimate of model confidence is crucial.
In this work, we propose a graph-based Bayesian convolutional neural network (CNN) for recovering the underlying CMB signal from multi-frequency temperature maps of the full-sky in a probabilistic manner. To avoid edge effects imposed by projecting spherical sky maps onto a two-dimensional plane, we employ HEALPix \cite{gorski2005healpix} - a graph representation that enables local operations while minimizing discretization error. We reformulate the state-of-the-art image-to-image network, U-Net \cite{ronneberger:unet15}, as a Bayesian framework that operates on the graph-based representation and provides probabilistic predictions. The proposed model shows promise in accurately recovering CMB temperature maps and quantifying data and model-dependent uncertainty associated with the prediction.

The contributions of this paper are (1) a novel approach to
CMB extraction that combines a graph-based representation
with a Bayesian U-Net CNN, (2) initial results on low-resolution
full-sky data with a comparison to existing methods for
CMB extraction that do not employ machine learning, (3) an analysis of uncertainty quantification calibration and quality, and (4) a discussion of the strengths and weaknesses of the proposed approach.

\section{Related Work} 
Sky component separation, i.e.,~extraction of 
primordial CMB anisotropy signals from multi-frequency
measurements of the microwave sky, is the key data processing step
to identify the cosmological model that best
describes the observed properties of the universe. The most
comprehensive comparison of component separation methods
applied 
to data collected by the Planck satellite was provided by the
\citet{planck2014results, planck2015resultsIX,
  planck2018results}.
These component separation methods range 
from linear data combination techniques to multiparameter Bayesian
model fitting \cite{rocha2019astro2020}. 
In this initial work, 
we compare our results to those 
derived using the simplest implementation of the Internal Linear
Combination (ILC) technique \cite{ilc1992}.

Recent work has explored employing neural networks as a data-driven approach to CMB extraction. Early approaches utilized fully-connected multilayer perceptrons, neglecting the inherent spatial correlation of the input maps \cite{norgaard2008foreground}. CNNs retain spatial correlations but are formulated for Euclidean input. There has been progress in training CNNs using flattened 2D patches of full and partial-sky maps \cite{wang2022recovering, casas2022cenn}; however, these 2D projections are prone to edge artifacts. DeepSphere \cite{perraudin:deepsphere19} presents an efficient spherical CNN that leverages the hierarchical HEALPix \cite{gorski2005healpix} graph representations for pooling and performs convolution in the spectral domain. A recent model \cite{petroff:cmb-cnn20} combines DeepSphere with a U-Net architecture \cite{ronneberger:unet15} for full-sky map CMB recovery. Additionally, this model employs concrete dropout for UQ. We employ a similar technique, but in \cite{petroff:cmb-cnn20} independent copies of the U-Net encoder are trained for each of the different frequency input maps, then features are combined and passed through the decoder. In our formulation, we treat the frequencies as channels of the same input and have a single encoder, making the model more scalable and preserving the cross-frequency correlation. 

Deep neural network uncertainty quantification (UQ) is an active line of research. The most straightforward technique to acquire such measures of model confidence is to train an ensemble of networks and consider the variance in predictions \cite{lakshminarayanan2017ensemble}. However, ensembles are computationally expensive and are not guaranteed to produce diverse output. The $M$-heads framework was proposed to remedy this by using one common network that produces $M$ different output \cite{lee2015m, rupprecht2017multiple}. Nevertheless, the $M$-heads method similarly scales poorly and lacks a principled theoretical foundation. 

A more principled approach is Bayesian deep learning, which uses Bayes' rule to learn a distribution over network weights, providing both aleatoric and epistemic uncertainty quantification. Aleatoric (or data-dependent) uncertainty is the irreducible uncertainty inherent in the input data (for example, noise resulting from measurement imprecision). It can be learned as a function of the input data via MAP inference by adjusting the loss function \cite{kendall2017uncertainty}. Epistemic (or model-dependent) uncertainty is the reducible uncertainty in model structure and parameters and is captured by placing a probability distribution over model weights. \cite{kendall2017uncertainty} showed that Monte Carlo (MC) dropout is a scalable, principled approach to approximate variational inference for acquiring epistemic uncertainty measures. Concrete dropout is a continuous relaxation of dropouts discrete masks that allows updating the optimization operation to automatically tune the dropout probabilities of a Bayesian network \cite{gal2017concrete}. 
An alternative approach to dropout combines a conditional variational autoencoder \cite{kingma2013auto} with a U-Net \cite{ronneberger:unet15} to learn a conditional density model over output conditioned on the input \cite{kohl2018probabilistic}. However, this approach does not differentiate between aleatoric and epistemic uncertainty and requires training auxiliary networks making it challenging to scale. 
For these reasons, we elect to use concrete dropout with heteroscedastic uncertainty loss in adapting our graph-based CNN to be Bayesian for UQ.

\section{Methods} 
Our model operates on full-sky maps, predicting the cleaned CMB from nine temperature observations (frequency bands in GHz: 30, 44, 70, 100, 143, 217, 353, 545, and 857). We use Bayesian graph-based CNN with U-Net architecture that preserves the spherical nature of the data and provides uncertainty reasoning.

\subsection{Graphical Representation and Convolution}
To address this, we perform convolutions in the spectral domain and represent sky maps as graphs using the Hierarchical Equal Area isoLatitude Pixelization (HEALPix \cite{gorski2005healpix}) scheme as is done in DeepSphere \cite{perraudin:deepsphere19}.
The Fourier transform of these HEALPix graph signals is the projection of the signal onto the set of eigenvectors of the graph Laplacian. In this work, we use the combinatorial Laplacian, which is defined as:
\begin{equation}
    \bL = \D - \A
    \label{eq:laplacian}
\end{equation}
where $\A$ is the weighted adjacency matrix of the graph and $\D$ is the diagonal degree matrix, which contains the sum of the weights of all the edges for each node on its diagonal.

Traditional graph convolutions are often performed in the spectral domain using
spherical harmonics, and the computational complexity of such convolution 
methods is $O({N_{pix}}^3)$. In convolutional neural networks (CNN), there 
are often tens to thousands of convolution operators, and each convolution 
operation is done in every forward pass during the training process. A
computational complexity of $O({N_{pix}}^3)$ convolution operation will almost
make the training process intractable. In this work, we use the efficient convolution operation proposed in 
~\cite{perraudin:deepsphere19}, which reduces 
the computational complexity of graph convolution to $O(N_{pix})$. This efficiency is obtained by representing each convolutional kernel, $h$, as a polynomial approximation using Chebyshev polynomials. Convolution for the CMB signal, $f$, is then defined as:
\begin{equation}
    h_\phi(\hat{\bL})f = \sum_{k=0}^{K} \theta_kT_k(\hat{\bL})f
    \label{eq:chebyshev_conv}
\end{equation}
where $T_k(\cdot)$ is the Chebyshev polynomial of degree $k$ defined by the 
recursive relation, $\hat{\bL}$ is the normalized form of the Laplacian matrix, 
and $\phi$ is the polynomial coefficients that are learned during the 
training process. 
In addition to the convolution operation, CNN architectures involve pooling and up-sampling layers. Pooling refers to an operation that summarizes (e.g., down-samples) a feature map by a pre-defined factor, and up-sampling is the inverse operation that expands a feature map by a given factor. This work uses maximum pooling and nearest neighbor up-sampling defined on HEALPix grids. The hierarchical tree structure of the HEALPix graph easily allows for such operations.

\subsection{Network Architecture and Loss}
\begin{figure}[ht!]
  \includegraphics[width=\linewidth]{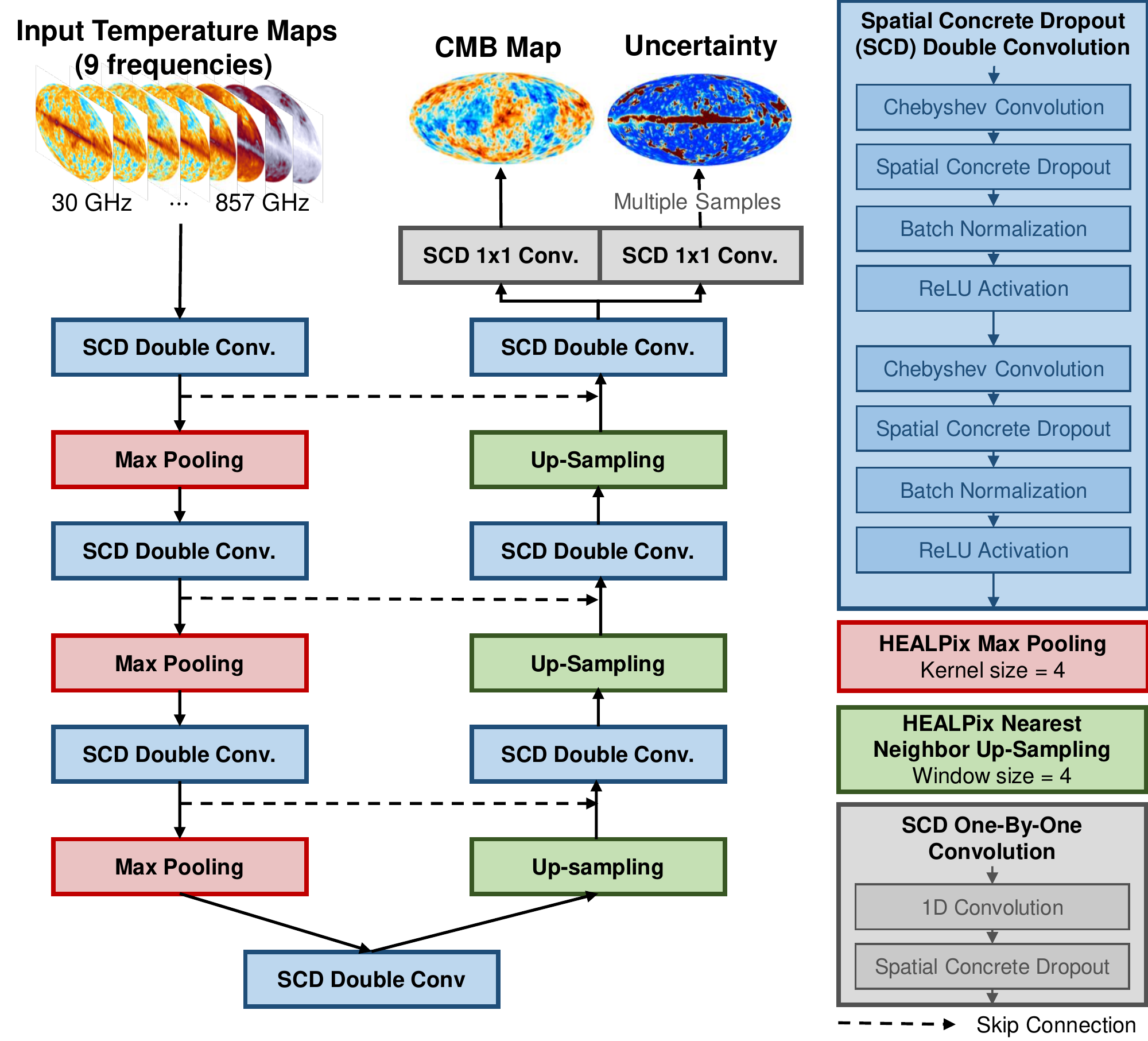}
  \caption{Model Architecture. The graph-based CNN has a U-Net based architecture. The down arrows represent the encoding path and the up arrows represent the encoding path.}
  \label{fig:cnn-archi}
\end{figure}

We utilize a Bayesian formulation of the U-Net architecture \cite{ronneberger:unet15} shown in Figure \ref{fig:cnn-archi}. It consists of an 
encoding and a decoding path. The encoding path consists of repeated operations 
of two Chebyshev convolutions with a Chebyshev polynomial order of 3, each followed
by a 1-dimensional batch normalization and a rectified linear unit (ReLU) \cite{relu}. 
After the two Chebyshev convolutions, the HEALPix-based max-pooling operation is 
applied to down-sample the feature maps by a factor of 4. Each depth of the
decoding path consists of a HEALPix-based nearest neighbor up-sampling operation, 
concatenation with the corresponding feature maps from the encoding path, two 
Chebyshev convolutions, each followed by 1-dimensional batch normalization and ReLU.

The network is formulated as Bayesian using concrete dropout for approximate variational inference and heteroscedastic aleatoric loss. Monte Carlo dropout was initially formulated as a technique for stochastic variational inference that uses a Bernoulli distribution for the approximate posterior distribution, $q_\theta(\bomega)$ (where $\theta$ are the variational parameters and $\bomega$ the network weights), adapting the model output stochastically \cite{kendall2017uncertainty}. However, obtaining well-calibrated uncertainty estimates requires a tedious grid search to tune dropout probabilities layer-wise. Spatial concrete dropout \cite{gal2017concrete} presents a continuous relaxation of Monte Carlo dropout’s discrete masks that allows updating the optimization operation to tune layer-wise dropout probabilities automatically. We apply spatial concrete dropout to every Chebyshev convolutional layer in the network in training and testing. We sample multiple predicted clean CMB maps with different dropout masks during testing and quantify the epistemic uncertainty as the variance across those maps.

The model captures aleatoric uncertainty by directly predicting a pixel-wise Gaussian distribution for the output CMB map. There are two one-by-one convolutional output layers; one predicts the mean, $\hat{\y}$ and the other the log variance, $\log(\hat{\sigma}^2)$ (note the log is predicted for numerical stability). The model is trained to predict both by maximizing the logarithm of the likelihood function under the observed data. The heteroscedastic aleatoric loss with concrete dropout is defined as:
\begin{equation}
\begin{split}
    \mathcal{L} = \frac{1}{D}\sum_{i=1}^D\big(\frac{1}{2}\hat{\sigma}_i^{-2}||\y_i - \hat{\y}_i||^2 + \frac{1}{2}\log{(\hat{\bsigma}_i^2)}\big) \\
    + KL(q_\theta(\bomega)||p(\bomega))
    \label{eq:loss}
\end{split}
\end{equation}
where $D$ is the number of pixels, $\y$ is the target CMB map, $\hat{\y}$ is the predicted CMB map, $\hat{\sigma}^2$ is the predicted variance, and the last term is the Kullback–Leibler (KL) regularization term \cite{kendall2017uncertainty, gal2017concrete}. The KL regularization term ensures that the approximate posterior $q_\theta(\bomega)$ does not deviate too far from the prior distribution $p(\bomega)$, where $p(\bomega)$ is selected to be standard Gaussian to allow for analytical evaluation.

Prediction uncertainty, denoted Var($\y$), is quantified as the sum of the epistemic and aleatoric uncertainty. Var($\y$) is acquired by sampling predictions with different concrete dropout masks $T$ times, then computing:
\begin{equation}
    \text{Var}(\y) \approx \frac{1}{T}\sum_{t=1}^T\hat{\y}_t^2 - \left(\frac{1}{T}\sum_{t=1}^T \hat{\y}_t\right)^2 + \frac{1}{2}\sum_{t=1}^T \hat{\bsigma}^2_t
    \label{eq:var}
\end{equation}
where $\hat{\y}_t, \hat{\bsigma}^2_t$ are the network predictions with the $t$-th weights sampled from the approximate posterior, $\bomega_t \sim q_\theta(\bomega)$. The first part of Equation \ref{eq:var}, the variance in the predicted means, is the epistemic uncertainty, and the second part, the average predicted variance, is the aleatoric uncertainty.


\subsection{Model Initialization}
As noted in the Bayesian deep learning literature, using the negative log-likelihood based loss in equation \ref{eq:loss} in conjunction with a gradient-based optimizer can lead to very poor but stable parameter estimates \cite{seitzer2022pitfalls}. This can result in poor predictive accuracy. We found that this pitfall can be avoided using a sensible transfer learning approach. We initialize the weights of our graph-based Bayesian CNN using a trained deterministic version of the CNN. The deterministic version has an identical architecture to the proposed model (Figure \ref{fig:cnn-archi}), except the predictive variance output layer and all spatial concrete dropout layers are removed. We train the deterministic network using mean square error loss:

\begin{equation}
    \mathcal{L}_{\text{deterministic}} =  \frac{1}{D}\sum_{i=1}^D||\y_i - \hat{\y}_i||^2 
    \label{eq:deterministic_loss}
\end{equation}
until it reaches convergence. We then initialize the Chebyshev convolutional weights in the Bayesian model using the weights learned by the deterministic model. This pre-training helps the Bayesian CNN achieve good predictive accuracy while learning to quantify uncertainty.

The output layer that predicts the variance, which is not pre-trained, is initialized to have small random values sampled from the Gaussian distribution, $\mathcal{N}(0,\num{1e-6})$. This initialization prevents the variance term from overshooting in early Bayesian training epochs. The layer-wise spatial concrete dropout probabilities are all initialized to be \num{1e-3}. This small initialization also makes model training more stable and helps account for any covariance shift caused by batch normalization in conjunction with dropout \cite{li2019disharmony}.

\section{Experimental Results} 

\subsection{Simulated Data}
We generate simulated sky observations following established practice
that combines three elements:

\begin{enumerate}
    \item $C_i$: A randomly generated realization of primordial CMB
      anisotropies (the target we seek to recover)
      from the distribution specified by Planck 2015 estimates
      \cite{planck2015resultsXIII} of the cosmological model
      parameters.
    \item $F$: Existing (fixed) maps, at each of the nine Planck frequency
      bands,
      of the composite foreground emission provided in
      the 2015 Planck data release \cite{planck2015resultsIX}. 
    \item $N_i$: Randomly generated, spatially modulated noise realizations
      consistent with Planck instrument sensitivity per frequency
      band.
\end{enumerate}

Each complete simulation is the sum of elements $C_i + F + N_i$.  We
reduced the angular resolution of the signal maps (CMB and
foregrounds) to FWHM=150 arcmin (unlike Planck, the same resolution
per frequency band, to simplify the analysis), and rendered the sky
maps at HEALPix resolution of NSIDE = 64. The resulting simulated
sky maps contain 49,152 pixels.

We generated 1,000 instances of noisy multi-frequency maps and clean
CMB pairs, each with a unique foreground sky map and unique noise realization. We split these instances into a training set, validation set, and held-out testing sets using a random 80\%, 10\%, 10\% split, resulting in a training set of 800, a validation set of 100, and a testing set of 100. The input temperature maps were normalized channel/frequency-wise to be zero-centered with a standard deviation of one. This standard input normalization improves model training. We found that applying normalization techniques to the target CMB maps did not improve model accuracy, so the model learns to predict un-scaled (not normalized) CMB maps. 

\subsection{Model Training}

\begin{figure}
  \includegraphics[width=\linewidth]{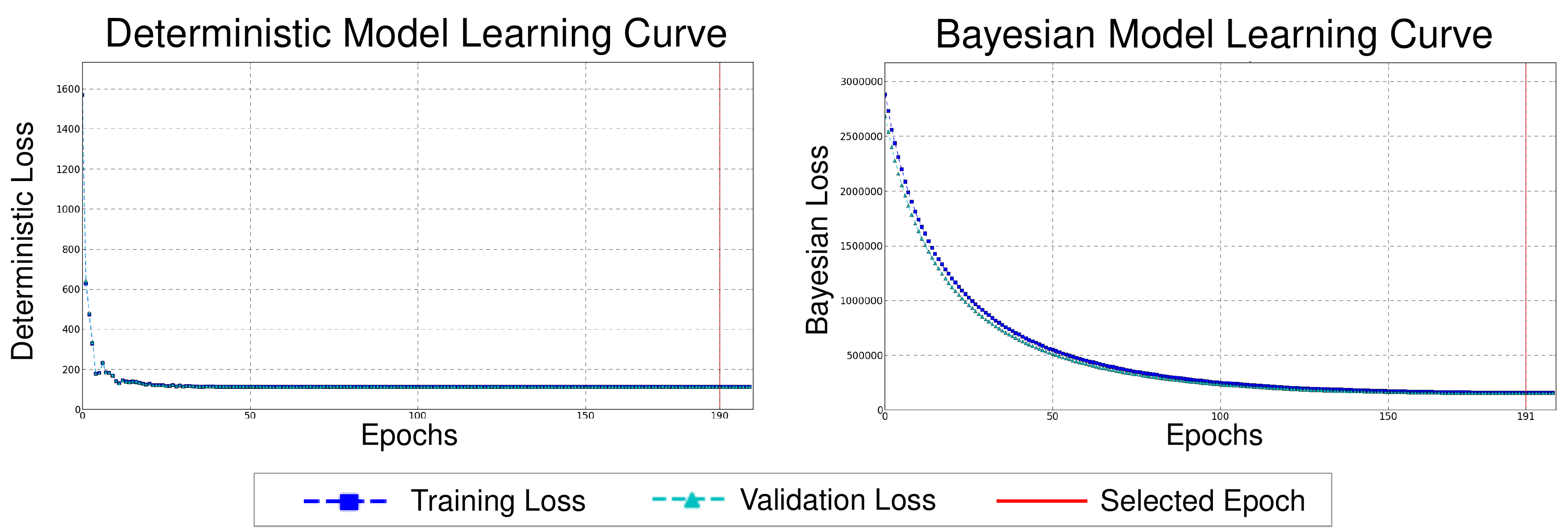}
  \caption{Model Training Learning Curves. The curves show training and validation loss over the epochs as well as the selected epoch to use for initialization or prediction. Note the validation loss is very close to the training loss, suggesting the models are not over-fitting to the training data.}
  \label{fig:curves}
\end{figure}

The model was implemented using PyTorch and trained on an NVIDIA Tesla M60 RAF GPU. The deterministic model used for weight initialization was trained with a learning rate of \num{1e-3} with SGD optimization and MSE loss (Equation \ref{eq:deterministic_loss}). A batch size of 10 was selected as this was the largest batch size the GPU memory would allow. 
In both the deterministic and Bayesian training, we saved the model after every training epoch and used the validation set to determine which model to use. We selected the epoch that minimized combined loss defined as: $0.8 * \text{(validation loss)} + 0.2 * \text{(train loss)}$. This metric allowed us to select the ideal stopping epoch to prevent over-fitting. The selected epoch for the deterministic model was 190, and it took 21 hours to train. 
The Bayesian model was found to train better with a lower learning rate (\num{1e-5}) and Adam optimization \cite{adam}. A length scale of \num{1e-4} was used for concrete dropout. The batch size was reduced to 7 in training the Bayesian model to account for the memory required by the additional concrete dropout parameters. The selected epoch for the Bayesian model was 191, and it took 51 hours to train. The combined training required 382 epochs and took about three days. 
The learning curves and selected model epochs are shown in Figure \ref{fig:curves}.

\subsection{Analysis of Results}
\label{section:results}

\begin{figure}[t!]
  \includegraphics[width=\linewidth]{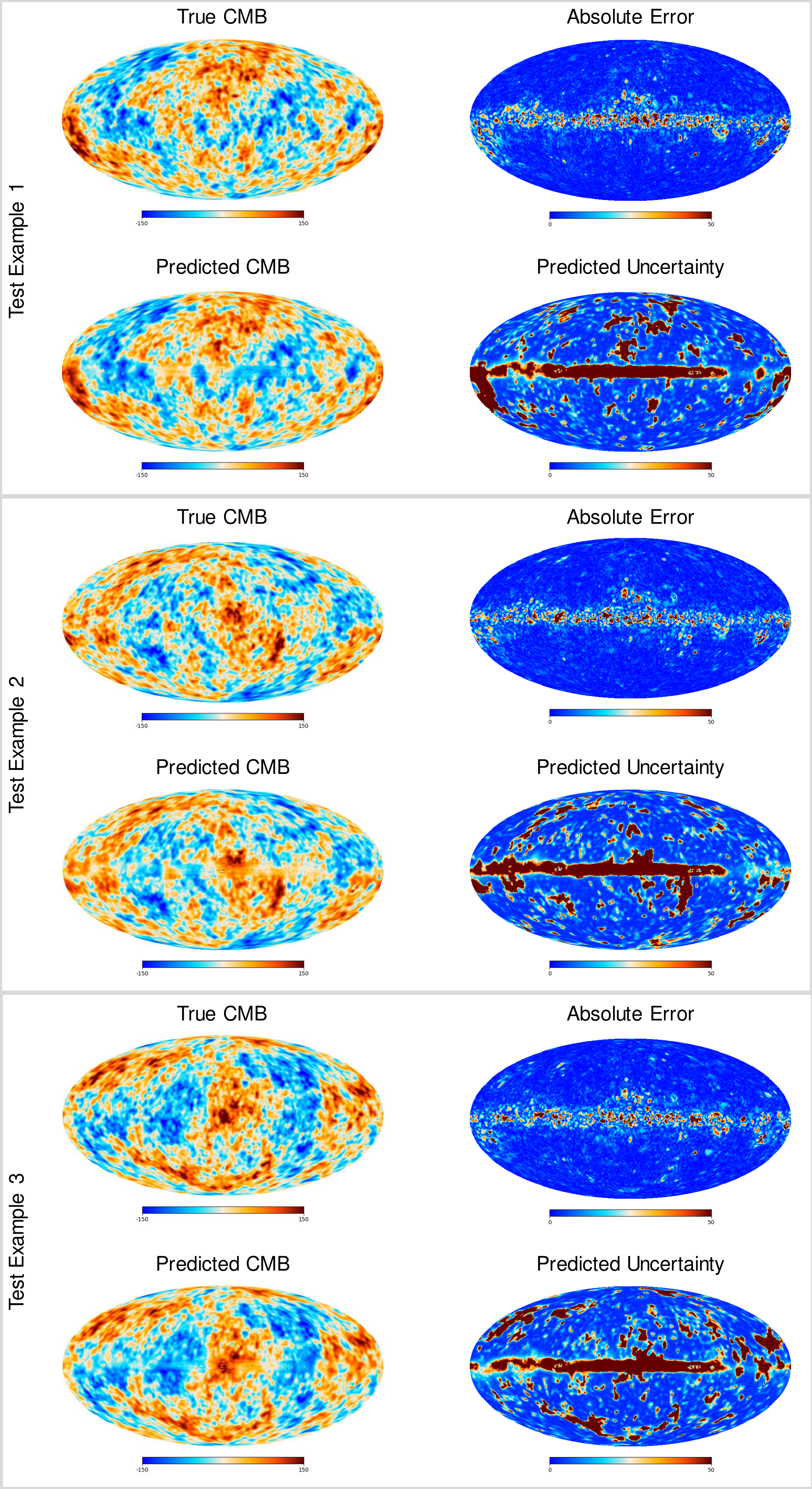}
  \caption{Model Prediction. An example from the test set is provided with the true and predicted CMB maps, absolute error or absolute difference between the true and predicted, and predicted uncertainty.}
  \label{fig:results}
\end{figure}

We provide qualitative and quantitative metrics for analyzing the prediction accuracy and uncertainty calibration on the simulated data. Figure \ref{fig:results} displays three examples from the test set. On the left, we can see the true CMB temperature map versus the one predicted by the model. On the right, we can see the absolute error, or the absolute difference between the true and predicted maps, and the predicted uncertainty. The predicted uncertainty is the standard deviation of the pixel-wise predicted distribution (or the square root of the variance given by Equation \ref{eq:var}). As expected, we can see the error and uncertainty are highest along the galactic plane. The predicted uncertainty should correlate with the error, and we observe that it does so, but it is over-estimated in some regions.

\begin{figure}[t!]
\begin{center}
  \includegraphics[width=\linewidth]{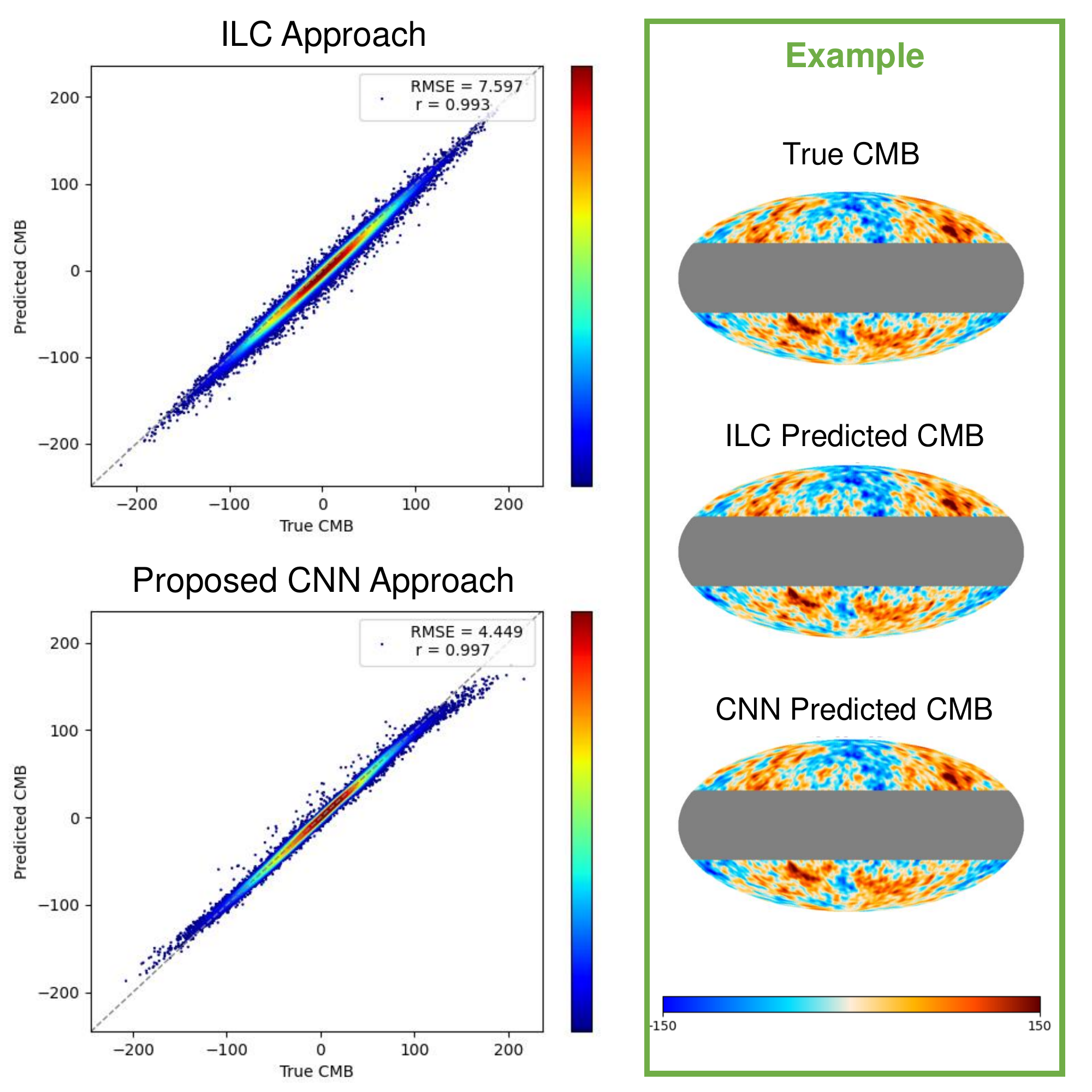}
  \caption{Pixel-wise Prediction Accuracy. Plots show correlation between true and predicted CMB pixel values (where values that lie on the diagonal are more accurately predicted) for the ILC and proposed method. An example from the test set is provided to illustrate the $\pm 30$ degree sky cut used in evaluation. }
  \label{fig:performance}
\end{center}
\end{figure}

To analyze prediction accuracy, we compare against the ILC method. We consider the overall root mean square error (RMSE) as well as the Pearson correlation coefficient between the true and predicted pixel values within a $\pm 30$ degree sky cut. On the test set, the ILC method predictions yield RMSE=7.597 and $r$=0.993. The predictions from the proposed CNN yield RMSE=4.449 and $r$=0.997, suggesting imporved accuracy over the ILC baseline, Figure \ref{fig:performance} illustrates how closely predicted pixel values correlate with the true CMB pixel values over the entire test for both methods. 
We note that the CNN predicted values are closer to the diagonal (i.e., more accurate) in general than the ILC values, they are farther from the diagonal for pixels with an actual CMB value of large magnitude. This can be seen in the slight "S" shape of the distribution in Figure \ref{fig:performance}, where the tails are under-estimated by the model. This issue is also manifested in the sky maps in Figure \ref{fig:results}. Regions of large temperature magnitude in the true CMB correlate to regions of slightly higher error in the absolute error maps and notably higher uncertainty in the predicted uncertainty maps. This indicates that the model has less confidence in predicting values of high magnitude, suggesting such regions are under-represented in the training data.

\begin{figure}[ht!]
\begin{center}
  \includegraphics[width=.8\linewidth]{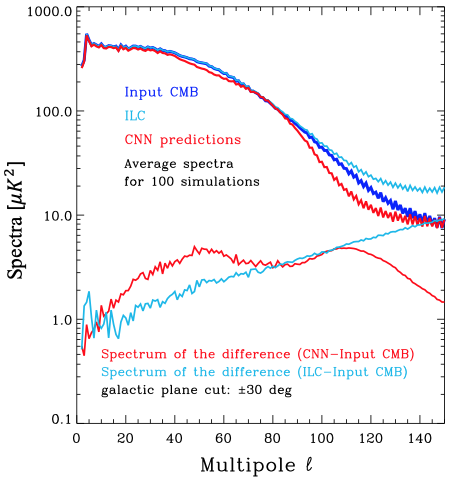}
  \caption{Spectral Comparison. The upper lines show the input CMB (true) spectra as well as those from the ILC and proposed CNN predictions. The lower lines show the spectra of difference maps between ILC and CNN predictions vs. input CMB (smaller is better).}
  \label{fig:spectral}
\end{center}
\end{figure}

In Figure \ref{fig:spectral}, we can see a comparison in the spectral domain between the test maps predicted by the proposed method and those generated from the ILC method. The spectra were computed on $\pm 30$ degree sky cuts and averaged over the 100 test realizations. In comparing the spectra from the true (input CMB) and the predictions from the ILC and proposed CNN, we can see that both approaches deviate. Notably, the ILC approach picks up noise at high $\ell$ values, and we see a drop in the power from the CNN predictions after $\approx 100 \ell$. This suggests that using the two approaches in tandem could provide better accuracy. 

\section{Conclusion and Future Work} 
We presented a Bayesian deep learning approach to CMB recovery via a probabilistic graph-based CNN. We demonstrated the efficacy of this approach in accurately predicting cleaned CMB maps using simulated low-resolution temperature maps of the full-sky. These results demonstrate the potential for using the proposed model to recover the CMB signal from Plank observations. Three major challenges need to be addressed to make this possible:
\begin{enumerate}
    \item \textbf{Improving accuracy in regions of high magnitude.} As previously noted, the model underestimates regions where the CMB temperature has a high magnitude and predicts high uncertainty in these areas. We find this issue in related works \cite{petroff:cmb-cnn20, wang2022recovering, casas2022cenn} as well, where there are structured differences in the true and recovered spectra, which cannot be accounted for by randomness or noise. Resolving this issue will improve prediction accuracy and uncertainty quantification calibration. We believe the root of this issue is that regions of high signal-to-noise are under-represented in the training data, resulting in low model confidence in predicting them. To resolve this, we plan to explore adapting the loss to weigh these regions to a greater extent as well as data augmentation techniques to reduce pixel-value imbalance in simulated data.
    \item \textbf{Scaling up to full-resolution sky maps.} Currently, we are using a HEALPix map resolution of NSIDE = 64 for faster experimentation and model training. However, the Planck observations provide a resolution of NSIDE 1024 to 4096. Acquiring the most precise estimation of the CMB requires using this full resolution. We plan to implement parallelization or distributed training across multiple GPUs to address the computational expense associated with scaling up to this high-resolution data.  
    \item \textbf{Predicting both temperature and polarization.} CMB anisotropies lie both in the temperature and polarization of the radiation. Currently, our approach and other existing works consider only fluctuations in temperature. Adapting the model to predict polarization amplitude and direction is required to recover the full CMB signal. Accurately predicting polarization will be more challenging than temperature because the polarization of the foreground emissions is an order of magnitude larger than the CMB. Integrating traditional techniques for CMB cleaning, such as Commander \cite{eriksen2008commander}, could improve our approach and help address this challenge. Such a technique could be applied first and provided as additional input to the graph-based CNN, which could then learn to refine the CMB cleaning further.
\end{enumerate}
Future work may also involve exploring different techniques for uncertainty quantification. The concrete dropout approach to variational approximate inference relies on the assumption that the weight prior is standard Gaussian. Methods that learn a prior, such as \cite{kohl2018probabilistic}, rather than assuming the prior, may provide more flexibility and thus better calibrated uncertainty quantification.


\paragraph{Path to Deployment.}

The graph-based Bayesian CNN approach proposed in this work has the potential to be used by current and future cosmology missions from NASA and ESA (the European Space Agency). Separating CMB signals from foreground emission and systematic noise provides the key input needed to constrain models of how the universe began and has evolved.  Currently the maps available on mission websites and archives lack pixel-wise uncertainties. Deployment would include integration into mission pipelines and publicly posting the CMB maps with uncertainties (e.g., \url{https://pla.esac.esa.int/pla/#maps}).  Two authors of this paper (Krzysztof M. Gorski and Gra\c{c}a Rocha) are cosmologists who worked as project scientists on ESA's Planck mission and can inform and guide integration.  All cosmologists could use these maps for their own further modeling.  The pixel-wise uncertainties will enable fine-grained identification of reliable and problematic sky regions as well as filtering for users who wish to exclude less confidently modeled regions.

\section{Acknowledgments}

We thank the Jet Propulsion Laboratory Research and Technology Development program for funding support and the Science Understanding from Data Science initiative leads, Susan Owen and Lukas Mandrake, for this project's continuous support. The High Performance Computing resources used in this investigation were provided by funding from the JPL Information and Technology Solutions Directorate. This research was carried out at the Jet Propulsion Laboratory, California Institute of Technology, under a contract with the National Aeronautics and Space Administration. Copyright 2022. All rights reserved.

\appendix

\bibliography{aaai23} 

\end{document}